# The EcoLexicon Semantic Sketch Grammar: from Knowledge Patterns to Word Sketches

**Pilar León-Araúz, Antonio San Martín**

University of Granada, University of Quebec in Trois-Rivières
Buensuceso, 11 18001 Granada (Spain), 3351, boul. des Forges, Trois-Rivières (Quebec, Canada)
pleon@ugr.es, antonio.san.martin.pizarro@uqtr.ca

**Abstract**

Many projects have applied knowledge patterns (KPs) to the retrieval of specialized information. Yet terminologists still rely on manual analysis of concordance lines to extract semantic information, since there are no user-friendly publicly available applications enabling them to find knowledge rich contexts (KRCs). To fill this void, we have created the KP-based EcoLexicon Semantic Sketch Grammar (ESSG) in the well-known corpus query system Sketch Engine. For the first time, the ESSG is now publicly available in Sketch Engine to query the EcoLexicon English Corpus. Additionally, reusing the ESSG in any English corpus uploaded by the user enables Sketch Engine to extract KRCs codifying generic-specific, part-whole, location, cause and function relations, because most of the KPs are domain-independent. The information is displayed in the form of summary lists (word sketches) containing the pairs of terms linked by a given semantic relation. This paper describes the process of building a KP-based sketch grammar with special focus on the last stage, namely, the evaluation with refinement purposes. We conducted an initial shallow precision and recall evaluation of the 64 English sketch grammar rules created so far for hyponymy, meronymy and causality. Precision was measured based on a random sample of concordances extracted from each word sketch type. Recall was assessed based on a random sample of concordances where known term pairs are found. The results are necessary for the improvement and refinement of the ESSG. The noise of false positives helped to further specify the rules, whereas the silence of false negatives allows us to find useful new patterns.

**Keywords:** EcoLexicon Semantic Sketch Grammar, knowledge patterns, sketch grammars, semantic relations, Sketch Engine

## 1. Introduction

Terminologists rely on corpus analysis for the extraction of conceptual information because most of the knowledge shared by experts is expressed in texts (Bourigault & Slodzian, 1999). For a long time, the only accessible way of analyzing corpus information for terminological work consisted in manually reading concordance lines. This is time-consuming and inefficient because for a given term a terminologist can be confronted with thousands of concordance lines, many of which may not carry any useful information for the terminologist.

Useful concordance lines for conceptual analysis are called knowledge-rich contexts (KRCs) (Meyer, 2001) and one of the most common approaches to find them is to search for knowledge patterns (KPs) in corpora. KPs are the linguistic and para-linguistic patterns that convey a specific semantic relation in real texts (Meyer, 2001). For instance, some of the simplest examples of generic-specific KPs are *x is a kind of y*, *As include Bs, Cs and Ds* (Meyer, 1994) and *comprise(s), consist(s), define(s), denote(s), designate(s), is/are, is/are called, is/are defined as, is/are known as* (Pearson, 1998).

KPs are considered one of the most reliable methods for the extraction of semantic relations (Auger & Barrière, 2008; Barrière, 2004; Bowker, 2003; Cimiano & Staab, 2005; Condamines, 2002; L'Homme & Marshman, 2006; Lafourcade & Ramadier, 2016; Lefever, Kauter, Hoste, Van de Kauter, & Hoste, 2014; Marshman, 2002, 2014; Marshman, Morgan, & Meyer, 2002). They have been applied in many terminology-related projects leading to the development of knowledge extraction tools, such as Caméléon (Aussenac-Gilles & Jacques, 2008) and TerminoWeb (Barrière & Agbago, 2006).

However, no user-friendly application allowing terminologists to find KRCs in their own corpora is publicly available. For this reason, in León-Araúz, San Martín & Faber (2016), we created a KP-based sketch grammar for Sketch Engine (Kilgarriff, Rychly, Smrz, & Tugwell, 2004) with the intention of allowing other users to extract KRCs through word sketches from their own corpora previously compiled with our grammar, which is mostly domain-independent.

Word sketches are defined as automatic corpus-derived summaries of a word's grammatical and collocational behavior (Kilgarriff et al., 2004). Rather than looking at an arbitrary window of text around the headword—as occurs in previous corpus tools—Sketch Engine is able to look for each grammatical relation that the word participates in (Kilgarriff et al., 2004). The default word sketches provided by Sketch Engine represent different relations, such as verb-object, modifiers or prepositional phrases. However, except for the recently implemented generic-specific word-sketches, they only represent linguistic relations. Figure 1 shows an example of three default word sketches in Sketch Engine.

| nouns modified by "bird" | | verbs with "bird" as object | | verbs with "bird" as subject | |
|---|---|---|---|---|---|
| | 24.17 | | 21.89 | | 21.15 |
| specie + | 17,353  9.09 | be + | 42,519  2.88 | be + | 88,426  3.38 |
| flu + | 11,801  9.82 | see + | 14,373  5.05 | have + | 22,998  3.40 |
| feeder + | 11,199  9.80 | kill + | 9,920  7.53 | fly + | 10,469  8.89 |
| watching + | 10,862  9.75 | have + | 7,084  1.77 | watch + | 8,554  8.19 |
| sanctuary + | 6,860  8.86 | watch + | 6,672  6.26 | sing + | 7,164  8.74 |
| life + | 6,168  5.24 | find + | 6,307  4.08 | do + | 7,039  3.59 |

Figure 1. Example of word sketches for *bird* in the English Web 2013 (enTenTen13) corpus

In León-Araúz, San Martín & Faber (2016), we developed 64 new sketch grammar rules focusing on the extraction of semantic relations, expanding the functionality of word sketches to the summarized representation of semantic behavior. This new sketch grammar for the English language includes some of the most common semantic relations used in the field of terminology: generic-specific, part-whole, location, cause and function. For the first time, this sketch grammar is now publicly available under the name of the EcoLexicon Semantic Sketch Grammar (ESSG). It is built in Sketch Engine to query the EcoLexicon English Corpus (see section 3.1), but users can also reuse it with any corpus following the instructions on <http://ecolexicon.ugr.es/essg>.

This paper describes the process of building a KP-based sketch grammar with special focus on the last stage, namely, the evaluation with refinement purposes. We conducted a shallow precision and recall evaluation of the 64 English sketch grammar rules created so far for hyponymy, meronymy and causality, which are an expansion and refinement of the ones presented in León-Araúz, San Martín & Faber (2016).

## 2. Building a KP-based sketch grammar

Although some authors (Marshman, 2004; Meyer, 2001) have inventoried patterns, they normally are a simplification of what is actually found in a corpus. For instance, when formalizing the pattern *is a type of* we should also take into account all of its possible variants. The verb *to be* may be in its plural form or substituted by a comma; if it is in the plural, various hyponyms will be enumerated to the left of the pattern; the verb *to be* may be preceded by a modal verb; the word *type* may be preceded by an adjective and an adverb; and it may be substituted by other synonyms such as *kind, sort, example, group*, etc. All of these possible variations must be accounted for when developing the grammar rules.

Corpus querying in Sketch Engine is based on an extension of the Corpus Query Language (CQL) formalism (Jakubíček, Kilgaiff, McCarthy, & Rychlý, 2010), allowing for the formalization of grammar patterns in the form of regular expressions combined with POS-tags. CQL expressions in Sketch Engine can be used as one-time queries (giving access to matching concordance lines) or stored in a sketch grammar, which will produce word sketches. For instance, if users query "[tag="JJ.*"] [lemma="energy"]" in SketchEngine, they will obtain all the concordances in which *energy* is preceded by an adjective in the corpus of their choice. For their part, sketch grammars are collections of CQL expressions that allow users to produce word sketches without any knowledge of the CQL formalism. A single word sketch may be the result of a combination of multiple long CQL expressions.

In the development of the ESSG we have considered different issues that are specific to each relation. For instance, there are certain patterns that always take the same form and order (e.g. *such as*), whereas others show such a diverse syntactic structure that the directionality of the pattern must also be accounted for. We also had to take into account the fact that a single sentence could produce more than one term pair because of the enumerations that are often found on each side of the pattern (e.g. *x, y, z and other types of w*). This entails performing greedy queries in order to allow any of the enumerated elements fill the target term. However, this may also cause endless noisy loops. Sometimes it is necessary to limit the number of possible words on each side of the pattern. In this sense, we observed that enumerations are more often found on the side of hyponyms, parts, and effects than on the side of hypernyms, wholes, and causes. Consequently, the loops were constrained accordingly in the latter case. Table 1 shows a summarized and simplified version of the patterns included for each semantic relation evaluated in this study (only a sample of 5 patterns per semantic relation for space reasons).

| |
|---|
| **Generic-specific:** HYPONYM ,(|:|is|belongs (to) (a|the|…) type|category|… of HYPERNYM // types|kinds|… of HYPERNYM include|are HYPERNYM // types|kinds|… of HYPERNYM range from (…) (to) HYPONYM // HYPERNYM (type|category|…) (,|() ranging (…) (to) HYPONYM // HYPERNYM types|categories|… include HYPONYM |
| **Part-whole:** WHOLE is comprised|composed|constituted (in part) of|by PART // WHOLE comprises PART // PART composes WHOLE // PART is|constitutes (a|the|…) part|component|… of WHOLE // WHOLE has|includes|possesses (…) part|component|… (,|() (:|such as|usually|namely|…) PART // WHOLE has|includes|possesses (a|the|…) fraction|amount|percent… of PART |
| **Cause:** CAUSE (is) responsible for EFFECT // CAUSE causes|produces|… EFFECT // CAUSE leads|contributes|gives (rise) to EFFECT // CAUSE-driven|-induced|-caused EFFECT // EFFECT (is) caused|produced|… by|because due (of|to) CAUSE |

Table 1: Simplified version of the patterns included in each grammar

By way of example, Table 2 shows the actual CQL representation of a generic-specific KP-based rule, followed by an explanation and three natural language examples of concordances matched with the grammar.

| 1:"N.*" [word=",|\("]? [tag="IN/that|WDT"]? "MD"* [lemma="be|,|\("] "RB.*"* [word="classified|categori.ed"] ([word="by"] [tag!="V.*"]+)? [word="in|into"] [tag!="V.*"]* [lemma="type|kind| example|group|class| sort|category|family|species|subtype| subfamily|subgroup| subclass|subcategory|subspecies"]? [tag!="V.*"]* 2:[tag="N.*" & lemma!="type|kind|example| group|class| sort|category|family|species|subtype|subfamily|subgroup| subclass|subcategory|subspecies"] | |
|---|---|
| 1:"N.*" | The hypernym is a noun. |
| [word=",|\("]? | An optional comma or bracket. |
| [tag="IN/that|WDT"]? | Optionally "that" or "which". |
| "MD"* | Any modal verb from zero to infinite times. |
| [lemma="be|,|\("] | Lemma "be" or a comma or a bracket. |
| "RB.*"* | Any adverb from zero to infinite times. |
| [word="classified|categori.ed"] | Classified, categorized, or categorized. |
| ([word="by"] [tag!="V.*"]+)? | Optionally, "by" followed by anything from one to infinite times that does not contain a verb. |
| [word="in|into"] | In or into. |
| [tag!="V.*"]* | Anything from zero to infinite times that does not contain a verb. |
| [lemma="type|kind| example|group|class|sort|category|family|species|subtype| subfamily|subgroup|subclass|subcategory|subspecies"]? | Optionally any of the lemmas "type", "kind", "example", "group", "class", "sort", "family", etc. |
| [tag!="V.*"]* | Anything from zero to infinite times that does not contain a verb. |
| 2:[tag="N.*" & lemma!="type|kind|example|group|class|sort|category|family|species|subtype|subfamily|subgroup|subclass|subcategory| | The hyponym is any noun other than "type", "kind", "example", "group", "class", "sort", "family", etc. |

| | |
|---|---|
| subspecies"] | |
| Stony-iron meteorites are classified into pallasites and mesosiderites. Modern reefs are classified into several geomorphic types: atoll, barrier, fringing, and patch. Littoral materials are classified by grain size in clay, silt, sand, gravel, cobble, and boulder. | |

Table 2. CQL representation of a generic-specific KP-based rule followed by its explanation

For the development of sketch grammar rules we followed the following methodology:

1. *Collection of KPs*: this first stage only includes the collection of patterns in plain English (no formalism or encoding language used).

-Patterns referenced by other authors.
-Patterns already known.
-Recursive method: term pairs linked by already known semantic relations are searched for to find new patterns. Then these patterns are used to find new term pairs, and so on.

2. *CQL encoding*: it consists of translating the KPs collected during the first stage into CQL sketch grammar rules.

-Splitting or lumping: some KPs collected in the first stage can be lumped into a single CQL sketch grammar rule, while others collected as a single KP need to be split.
-Addition of adverbs, punctuation, modal verbs, relative phrases, adjectives, determiners, etc.

3. *Enrichment and refining*: CQL rules are enriched and refined trying to keep the balance between noise and silence.

-Enrichment: Testing the CQL rules with additional optional elements to spot new variations of the pattern (for instance, the possibility of an adverb in a place where it was not previously accounted for).
-Refining: Detection of erroneous concordance lines obtained with the CQL rules. Analysis of the source of the error, and determination of whether it is appropriate to change the CQL rule.

4. *Evaluation*: this includes a precision and recall analysis, which is described in section 3.2. After the evaluation, the enrichment and refining step is repeated to include the new patterns and modifications that the analysis of noise and silence has proved necessary.

## 3. Evaluation of the ESSG

### 3.1 EcoLexicon English Corpus

For evaluating the ESSG, we applied them to the EcoLexicon English Corpus (EEC). The EEC is a 23.1-million-word corpus of contemporary environmental texts compiled by the LexiCon Research Group for the development of the environmental terminological knowledge base EcoLexicon (Faber & Buendía, 2014; Faber, León-Araúz, & Reimerink, 2016; San Martín et al., 2017)[1]. It can be queried within the knowledge base, but the corpus has also recently been made freely available in Sketch Engine Open Corpora[2]. Each text in the EEC is tagged according to a set of XML-based metadata. This allows constraining corpus queries based on pragmatic factors such domain, user, geographic variant, genre, editor, year and country of publication.

The EEC is tagged with the Penn Treebank tagset (TreeTagger version) ver. 3.3, which allows for more fine-grained queries in CQL. It employs the default sketch grammar for English in combination with the ESSG In this way, word sketches in the EEC incorporate automatic corpus-derived summaries of a concept's semantic relations (Figure 2). Thus, the aim of our sketch grammar is twofold: (1) offering semantic word sketches in our freely available EEC, (2) and providing other users (i.e. terminologists) with the possibility of reusing it in their own corpora.

| "mineral" is the generic of… | | "mineral" is part of… | | "mineral" is a type of… | | "mineral" has part… | |
|---|---|---|---|---|---|---|---|
| | 1,909 19.03 | | 985 9.82 | | 652 6.50 | | 472 4.71 |
| quartz | 60 9.86 | rock + | 144 10.66 | find | 6 8.14 | silicon | 22 10.20 |
| feldspar | 38 9.26 | soil | 30 8.77 | resource | 20 8.02 | oxygen | 26 9.56 |
| iron | 44 9.12 | magma | 12 8.54 | rock | 20 7.89 | carbonate | 18 9.39 |
| gold | 36 9.09 | melt | 10 8.34 | earth | 14 7.79 | magnesium | 14 9.28 |
| carbonate | 36 9.06 | silt | 10 8.26 | substance | 12 7.67 | calcium | 16 9.26 |
| mica | 32 9.02 | crust | 12 8.19 | constituent | 6 7.63 | iron | 16 9.25 |
| calcite | 28 8.85 | limestone | 10 8.12 | way | 6 7.59 | co3 | 8 9.00 |
| copper | 28 8.65 | peridotite | 8 8.01 | material | 26 7.42 | anion | 8 8.98 |

Figure 2. Word sketches of *mineral* in the EEC extracted with the ESSG

### 3.2 Precision and recall metrics

Precision is measured on a random sample of concordances of one of the terms that has most frequently been annotated as part of each word sketch. This leads to the identification of false positives and the analysis of their causes, which results in the refinement of sketch grammar rules. Given that at this stage the goal of the evaluation was to use the results to improve our sketch grammar before objectively assessing their global efficiency as knowledge extraction devices, we chose to analyze only the results of one particular term. This allowed us to reduce the workload of the evaluation process. Moreover, since sketch grammars are conceived for the compilation of word sketches that users might find interesting to look at, the keyword is chosen based on a term susceptible to being queried, avoiding, for instance, top-level concepts.

Recall, in turn, is measured on a random sample of concordances where the most frequent term pair is found, enriching the grammar rules through the identification of new useful KPs based on the false negatives encountered. The recall analysis is based on a particular term pair because that makes having a sample of manually curated positive concordances viable. The steps for each measure are as follows. Steps from 1 to 3 are common to both, with the only difference that for the precision analysis we select one particular term and for the recall analysis we select a particular term pair.

1. All concordances where each relation has been annotated are retrieved. For example, for the hyponymic relation the query [ws(".*-n","\"%w\" is a type of...",".*-n")] provides all the results where hypernyms and hyponyms (variables 1 and 2) have been annotated while compiling the corpus.

---

[1] ecolexicon.ugr.es/

[2] the.sketchengine.co.uk/open

2. The results are sorted by frequency with Sketch Engine's functionality Node form, showing the terms/term pairs that have most frequently annotated as one/both of the variables.

3. One of the most frequently annotated terms/term pairs is selected avoiding top-level concepts (i.e. *factor*, *parameter*) and terms that usually act as a modifier. Given the fact that users will query word sketches to find meaningful term pairs, we considered that broad top-level concepts are markedly less susceptible of being searched and thus we did not select them. Terms usually acting as modifiers were avoided as well since sketch grammars can only find single-word terms as variables for the moment.

*Precision:*

4. A sample of 1000 randomized concordances of the selected term is analyzed in order to quantify true and false positives.

5. The causes of false positives are analyzed and further constraints are defined in order to refine the grammar rules.

*Recall:*

4. A new query is performed in order to find all contexts where the pair occurs. For instance, the query (meet [lemma="wind"] [lemma="wave"] -15 15) within  provides all contexts within the same sentence where *wind* and *wave* are found in a word span of ±15.

5. From a randomized sample of 1000 concordances, we manually select all explicit occurrences of the relation in question, whether it is through KPs covered by the grammar or not.

6. A subcorpus is created based on the selected concordances, where we again perform the query in step 1 and then apply a negative filter. In this way, all concordances filtered are the ones that have not been identified by the grammar (false negatives).

7. The causes of false negatives are analyzed and further patterns are found in order to enrich the grammar.

## 4. Evaluation Results and Enhancement of the ESSG

The keywords selected for the precision analysis are: *species*, as a hypernym; *rock*, as a part; and *erosion*, as an effect. The term pairs selected for the recall analysis are: *breakwater-structure*, for hyponymy; *mineral-rock*, for meronymy; and *wind-wave* for causality. The concordances were extracted from the EEC.

As shown in Figure 3, hyponymic rules for *species* as a hypernym are 69.5% precise, whereas meronymic and causality rules scored 71.4% and 55.2% respectively. Recall was 45.2% for the hyponymic pair, 65% for the meronymic pair and 60% for the causality pair. Meronymic rules are thus the ones that perform better in terms of both precision and recall. Causal rules score better results for recall than for precision.

Considering that Sketch Engine only displays statistically relevant word sketches, the precision rate reached by the ESSG seems good enough to get reasonable results when users query the corpus to get semantic word sketches, such as those shown in Figure 1. The study of false positives (Section 4.1) and false negatives (Section 4.2) will contribute to the improvement and refinement of the grammar.

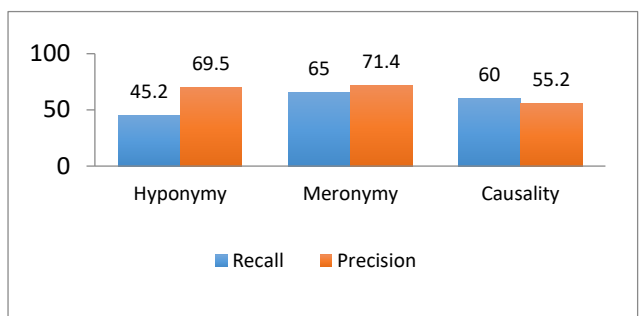

Figure 3. Precision and recall of hyponymic, meronymic and causality sketch grammar rules

### 4.1 Precision: analyzing false positives

Some FPs are caused by inherent limitations of using KP-based extraction of semantic relations with word sketches. Thus, we currently have no way of avoiding them.

1. POS-tagger mistake (mainly, tagging verbs as nouns).
   *…other species, especially those growing in natural ecosystems…*
2. Polysemous keywords: word sketches are unable to perform word sense disambiguation. Consequently, if the keyword is polysemous, the word sketch will show the results of all the senses combined (e.g. *species* as the hypernym of chemicals).
   *…scavenge the reactive oxygen species, including superoxide and hydrogen peroxide…*
3. The cause is a clause, not a noun.
   *They also trampled and over-grazed land, causing erosion and…*
4. Error induced by anaphora.
   *… a Dimilin–propanil mixture on these and other nontarget aquatic species.*
5. A correct relation is detected by mistake.
   *For Caulerpa taxifolia, the other Mediterranean invasive Caulerpa species, a decrease in specialist grazers such as Mullus surmuletus…*
6. The relation is only correct if transitivity is applied.
   *The basement to the arc is made up of at least 3000 m of Triassic (about 240 Ma) sedimentary rock…*

There are other types of FP that can be completely or partially avoided by refining our sketch grammar:

7. The detected hyponym/part/cause is a general word used as part of the pattern itself (i.e. *type*, *part*, *cause*).
   *More than a dozen Queensland frog species, especially the stream-dwelling types…*

All hyponymic grammar rules could be refined by negating for both variables (i.e. hyponym and hypernym)

the words that are used as anchoring words in the patterns. For instance, the rule that caused this FP could be transformed as follows (changes are highlighted in red): 1: [tag="N.*" & lemma!="type|kind|example|group|class|sort|category|family|species|subtype|subfamily|subgroup|subclass|subcategory|subspecies"] [word=",|\("] [word="especially|primarily|namely|usually|typically|characteristically|generally|mainly|particularly|chiefly|mostly|principally"] [tag!="V.*|IN"]* 2: [tag="N.*" & lemma!="type|kind|example|group|class|sort|category|family|species|subtype|subfamily|subgroup|subclass|subcategory|subspecies"]

8. Wrong detection of noun phrase.
   *...populations of the same or closely related <u>species</u> by a physical barrier such as a large <u>river</u> or...*
9. Error induced by the fact that the right elements of the pair are separated by too many words.
   *Streaming winds and following seas toppled expensive summer cottages into the surf, scrubbed the wooden-shingled roofs from quaint boutiques and <u>restaurants</u>, and caused extensive dune <u>erosion</u>.*

The solution in these cases (8 and 9) mostly lies in constraining very long loops. For instance, as mentioned above, in order to find enumerations of different terms at the left and right of the patterns we included broad loops such as [tag!="V.*"] (any word not being a verb). Instead, we should specify how enumerations are usually codified. With [tag="DT|RB.*|JJ.*|N.*" |word="and|or|,|;"]{0,10} we could gain in precision. However, an analysis will be needed to determine whether we would lose recall.

10. Error induced by a relative clause.
    *Ice sheets that form during <u>glaciations</u> cause <u>erosion</u>...*
    In this case, introducing relative clause markers (i.e. *that*, *which*) as a compulsory element between variables 1 and 2 would enhance causal grammar rules.
11. Error induced by negative sentences.
    *...<u>water</u> to enter into the test section from the head tank without causing immediate <u>erosion</u> and...*
    Constraints should be added to easily filter out these matches, adding a list of negative words (*never*, *without*, *no*, *not*, etc.) to all grammar rules.

### 4.2   Recall: analyzing false negatives

As a result of the recall analysis, the following patterns will be updated (changes are highlighted in gray):

- HYPERNYM ,|( such as|like (a|the|…) HYPONYM
- (a|the|one|two|some|…) part|component|building block… of WHOLE (is) called|referred… (to) (as) PART
- (a|the|one|two|some|…) part|component|building block… of WHOLE is PART
- PART (,|() (a|the|…) part|component|building block… of WHOLE
- PART (is) contained|present in WHOLE
- PART composes|constitutes|makes (up) WHOLE
- PART is|constitutes (a|the|…) part|component|building block… of WHOLE
- CAUSE causes|produces|creates… EFFECT
- EFFECT (is) caused|produced|created… by|because|due (of|to) CAUSE

The following are new patterns encountered during recall analysis, some of which might be integrated into existing patterns:

- major HYPERNYM is|include HYPONYM
- HYPERNYM (is) used as HYPONYM
- HYPERNYM serve|act as HYPONYM
- HYPERNYM ,|( e.g. |viz (a) HYPONYM)
- HYPONYM or any ADJ and ADJ HYPERNYM
- HYPERNYM (HYPONYM…
- HYPERNYM: HYPONYM
- HYPERNYM, these being HYPONYM
- WHOLE (is) rich in PART
- PART-rich WHOLE
- WHOLE is an aggregate of PART
- WHOLE and|or its part|component|… PART
- PART in|within WHOLE
- WHOLE with a proportion of PART
- percentage of WHOLE in PART
- EFFECT is the product of CAUSE
- CAUSE acts as generator of EFFECT
- CAUSE acts to cause|produce|create… EFFECT
- CAUSE contributes to the generation of EFFECT
- EFFECT generation by|due to CAUSE
- generation of EFFECT by|due to CAUSE

### 5.   Conclusions and future work

The evaluation performed on the ESSG has shown that even a shallow precision and recall analysis is an efficient way of detecting ways of refining and enriching the sketch grammar. Additionally, although the ultimate purpose of the evaluation was not to assess the global performance of the ESSG, the results suggest that the combination of word sketches with KPs has the potential of providing a reliable user-friendly method for the extraction of semantic relations in specialized corpora. Nonetheless, the evaluation indicates as well that there is still room for improvement as far as the level of precision and recall is concerned.

In future work, a larger evaluation study of all of our refined sketch grammar rules will be performed. This will include the study of each relation with no keyword limitations, the assessment of each pattern separately and the evaluation of word sketch precision for multiple term types. In addition to incorporating the improvements revealed by the precision and recall evaluations, the ESSG in the EEC will be enhanced by the inclusion of multiword terms based on those contained in the knowledge base EcoLexicon (by means of corpus annotation) and new collocation rules.

### 6.   Acknowledgements

This research was carried out as part of project FFI2017-89127-P, Translation-oriented Terminology Tools for Environmental Texts (TOTEM), funded by the Spanish Ministry of Economy and Competitiveness.

### 7.   Bibliographical References

## 8. Language Resource References